\documentclass{article}
\usepackage{iclr2026_conference}
\usepackage{times}
\usepackage{graphicx}
\usepackage{amsmath, amssymb}
\usepackage{natbib}
\usepackage{booktabs}
\usepackage{multirow, makecell}
\usepackage[normalem]{ulem} 
\usepackage[dvipsnames]{xcolor}
\usepackage{subcaption}

\usepackage{hyperref}
\hypersetup{
    colorlinks=true,
    linkcolor=BrickRed,
    citecolor=NavyBlue,
    urlcolor=NavyBlue
}
\usepackage{url}
\iclrfinalcopy

\title{\centering Understanding Representation Gaps Across Scales in Tropical Tree Species Classification from Drone Imagery}
\author{
  \begin{minipage}{\textwidth}
    \centering
    Sulagna Saha$^{1,2}$, Arthur Ouaknine$^{1,2}$, Etienne Laliberté$^{3,1}$, Carol Altimas$^{3,1}$ \\
    Evan M. Gora$^{4,5}$, Adriane Esquivel Muelbert$^{6,7}$, Ian R. McGregor$^{4}$ \\
    Cesar Gutierrez$^{4}$, Vanessa Rubio$^{4}$, David Rolnick$^{1,2}$ \\
    \vspace{0.8em}
    $^1$Mila -- Quebec AI Institute \quad $^2$McGill University \\
    $^3$Université de Montréal \quad $^4$Cary Institute of Ecosystem Studies \\
    \quad $^5$Smithsonian Tropical Research Institute \\
    \quad $^6$Department of Plant Sciences, University of Cambridge \\
    \quad $^7$Universidade do Estado do Mato Grosso (UNEMAT) \\
  \end{minipage}
}
\begin{document}
\maketitle

\begin{abstract}
\small

Accurate classification of tropical tree species from unoccupied aerial vehicle (UAV) imagery remains challenging due to high species diversity and strong visual similarity among species at typical image resolutions (centimeters per pixel). In contrast, models trained on close-up citizen science photographs captured with smartphones achieve strong plant species classification performance. Recent advances in UAV data acquisition now enable the collection of close-up images that are spatially registered with crown-view aerial imagery and approach the level of visual detail found in smartphone photographs, with the trade-off that such high-resolution photos cannot be acquired for many trees. In this work, we evaluate the performance of existing methods using paired crown-view and close-up UAV imagery collected in a species-rich tropical forest. Through fine-tuning experiments, we quantify the performance gap between vision foundation models and in-domain generalist plant recognition models across both image types (high-resolution close-up versus coarser-resolution crown-view imagery). We show that classification performance is consistently higher on close-up images (77.9\%) on single date than on crown-view aerial imagery (74.3\%) even after aggregating over 16 dates, and that this performance gap widens for rare species. Finally, we propose that self-supervised representation alignment across these two spatial scales offers a promising approach for integrating fine-grained visual information into canopy-level species classification models. Leveraging high-resolution close-up UAV imagery to enhance canopy-level species classification could substantially improve large-scale monitoring of tropical forest biodiversity.
\end{abstract}

\section{Introduction}

\begin{figure}[t]
    \centering
    \makebox[\textwidth][c]{%
        \includegraphics[width=\textwidth]{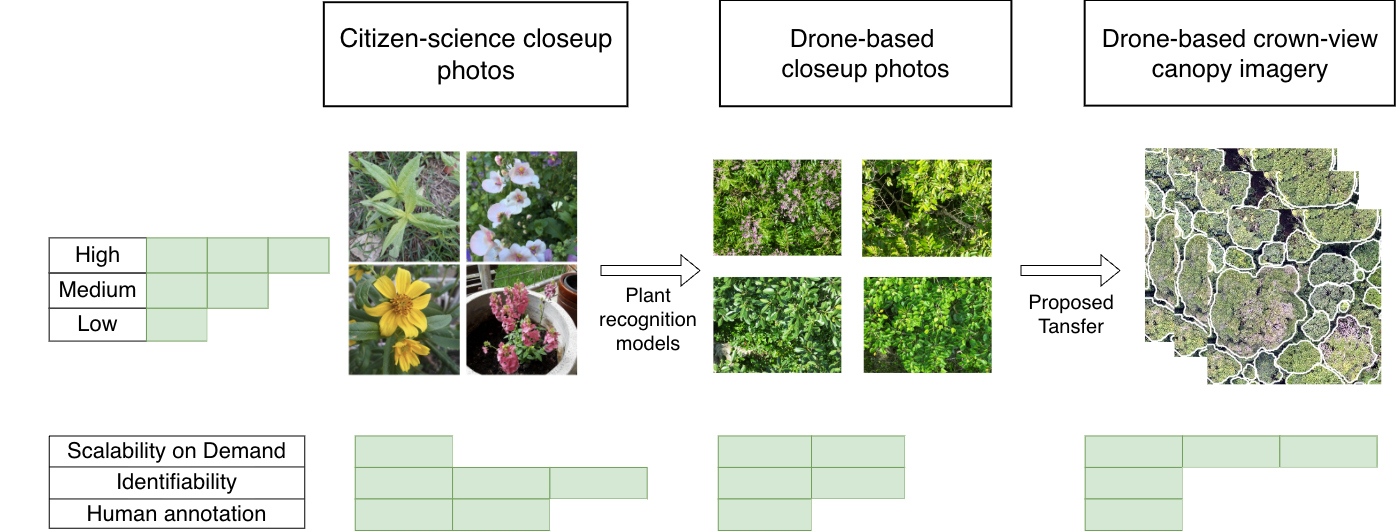}
    }

   \caption{From left to right, we show citizen-science close-up photographs \citep{plntnet_observations_2025}, drone-acquired close-up images, and drone-based crown-view canopy imagery. Citizen-science images offer high species identifiability and are easily annotated by humans but are weakly scalable on demand. Drone-based close-ups reduce annotation effort and improve scalability but introduce a domain shift relative to citizen data. Crown-view canopy imagery is the most scalable modality for large-area monitoring, yet lacks fine-grained botanical cues.}
    \label{fig:multiscale_gap}

\end{figure}

Tropical forests are the most biodiverse terrestrial ecosystems on Earth, harboring more than half of all tree species while occupying only about 10\% of the global land area \citep{beechGlobalTreeSearchFirstComplete2017}. Large canopy trees are particularly important because of their disproportionate contributions to carbon storage and ecosystem functioning \citep{slik_large_2013}. Despite their ecological significance, we know remarkably little about tropical canopy tree species. \citep{esquivelmuelbert_compositional_2019}. Individual tree species often exhibit distinct responses to environmental change, making accurate canopy-level biodiversity monitoring both essential and challenging \citep{araujo_integrating_2020}. Traditional ground-based surveys are costly, labor-intensive, and difficult to scale across large or remote regions \citep{forestplotsnet_taking_2021}, while satellite imagery frequently lacks the spatial resolution required for reliable species classification \citep{phillips_sensing_2023}. High-resolution crown-view RGB UAV imagery offers a promising, scalable alternative; however, annotating such data requires expert botanical knowledge and typically results in severe class imbalance, particularly for rare species \citep{schiefer_mapping_2020}.
Earlier work has shown that species-level tree classification from UAV imagery is feasible when high-resolution RGB data and accurate individual crown delineation are available \citep{kattenborn_review_2021}, with strong performance reported primarily in temperate forests, plantations, and other low-diversity systems where species exhibit pronounced morphological differences and labeled data are more abundant \citep{ferreira_identification_2023}.
However, recent studies demonstrate that classification accuracy degrades sharply as species richness increases and inter-species visual differences become more subtle, even with accurate crown segmentation \citep{teng_bringing_2025, nasiri_using_2025}. This challenge is exacerbated even more in tropical forests, which exhibit extreme species richness. \citep{cooperConsistentPatternsCommon2024}. High-resolution close-up imagery, acquired through citizen science platforms \citep{boone_using_2019, garcin_plntnet-300k_2021} or targeted drone flights \citep{laliberte_seeing_2025, zhang2016seeing} can help as they capture fine-grained characteristics such as leaf shape and arrangement, or flowers and fruits, that enable botanists to reliably identify species (Fig. \ref{fig:multiscale_gap}). Modern plant recognition models such as Pl@ntNet are predominantly trained on diverse citizen science close-up photographs \citep{Lefort2026}, which differ significantly from crown-view RGB UAV imagery in terms of spatial resolution, acquisition geometry, viewpoint, illumination, and background. Recent studies show how these models can be leveraged to produce meaningful, species-relevant representations when applied to UAV imagery \citep{soltani_transfer_2022}, but this has not been explored in species-rich tropical forests.


Species-discriminative visual cues are often ambiguous in crown-view canopy imagery acquired at centimeter-scale resolutions \citep{schiefer_mapping_2020, cloutier_influence_2024}. Recent drone-based workflows now enable the rapid and low-cost acquisition of close-up canopy photographs at sub-millimeter resolution (approximately 0.4~mm) \citep{laliberte_seeing_2025}, substantially narrowing the gap between conventional crown-view UAV imagery and close-up citizen science photographs (Fig.~\ref{fig:multiscale_gap}). In this study, we leverage drone-acquired close-up imagery to transfer fine-grained species information from plant recognition models to canopy-level classification. Our results show that plant recognition models generalize to both close-up and crown-view drone imagery for tree species classification, even under severe label scarcity and long-tailed class distributions. The observed performance gap between these spatial scales highlights opportunities for cross-scale representation alignment as a scalable pathway toward improved biodiversity monitoring in species-rich tropical forests.

\section{Dataset}
\label{sec:dataset}

We conduct experiments using high-resolution RGB drone imagery collected over Barro Colorado Island (BCI) during 2024--2025 \citep{sulagna_saha_2026}. The dataset comprises monthly whole-island orthomosaics captured at around 4 cm GSD. Such high-resolution UAV imagery is critical for reliable individual tree crown detection \citep{baudchon_selvabox_2026}, delineation \citep{duguay_selvamask_2026}, and species-level classification in structurally complex tropical forests. 
To enable individual-based analysis, we segment the RGB orthomosaics into tree-level crown-view polygons using \texttt{CanopyRS}, an automated canopy segmentation pipeline designed for high-resolution aerial imagery \citep{baudchon_selvabox_2026, duguay_selvamask_2026}. Using the \texttt{geodataset} v0.2.21 \footnote{\url{https://github.com/hugobaudchon/geodataset}} Python package, we extract 512~$\times$~512 RGB image tiles centered on each crown polygon. Pixels outside the polygon boundary are masked with black values to prevent background leakage and enforce crown-focused learning. The dataset time series enables observation of each individual tree across up to 16 monthly snapshots. These multi-temporal observations capture phenological and illumination variability while maintaining consistent spatial alignment. We also leverage a limited set of close-up images acquired during targeted drone missions designed to support taxonomic identification \citep{sulagna_saha_2026}.


In total, close-up imagery is available for 5,302 crown polygons, of which 1,999 polygons have species labels (annotated by expert from tropical regions). We restrict our classification experiments to 84 species that have at least 1 labeled individual for training available across the dataset. This results in a highly imbalanced class distribution dominated by rare species. For labeled data, crown polygons are randomly assigned to training (70\%), validation (15\%), and test (15\%) splits, resulting in 1,385 training, 288 validation, and 326 test labeled polygons (1,999 total). All temporal observations of a given tree are confined to the same split. We adopt a random polygon-level split rather than a geospatial split for two reasons. First, the dataset exhibits a large number of species with highly imbalanced frequencies, and enforcing spatial separation would substantially reduce rare-species representation in validation and test sets, leading to unstable and uninformative performance estimates. Second, all model inputs are derived from individually segmented crown polygons, and we explicitly remove all pixels outside each segmentation mask, removing pixel-level information leakage across splits. Close-up images inherit the split assignment of their corresponding crown polygon when labels are available. 

\section{Experiments \& Results}

We evaluate a diverse set of vision models commonly used in ecological image recognition. These include \textbf{ResNet-50} \citep{he2015deepresiduallearningimage} as a supervised convolutional baseline, \textbf{DINOv3} \citep{simoni2025dinov3} as a self-supervised vision transformer pretrained on large-scale image collections, \textbf{BioCLIP2} \citep{gu_bioclip_2025} as a biologically informed vision-language model, and \textbf{Pl@ntNet} \citep{lefort_cooperative_2026} as a plant recognition model based on DINOv2, pre-trained primarily on millions of close-up botanical photographs (using up-to-date weights for the production Pl@ntNet pre-trained model). The model specifications are mentioned in Table~\ref{tab:model_specs}.
For crown-view canopy images, comprising 16 temporal observations per tree, we evaluate two settings  (Table~\ref{tab:topview}): \textit{individual-image}, where each date is treated as an independent sample, and \textit{soft-voting}, where predicted class probabilities are averaged across the 16 temporal samples. 

\begin{table}[t]
\centering
\caption{Performance on crown-view canopy images. Models are fine-tuned on labeled canopy data across 16 acquisition periods. We report both individual-image predictions without taking time into account and crown-level soft voting aggregation.}
\label{tab:topview}
\begin{tabular}{llcccccccc}
\toprule
\textbf{\makecell{Evaluation\\mode}} &
\textbf{Model} &
\multicolumn{3}{c}{\textbf{Accuracy (\%)}} &
\multicolumn{3}{c}{\textbf{F1 Score}} \\
\cmidrule(lr){3-5} \cmidrule(lr){6-8}
& & \textbf{Top-1} & \textbf{Top-3} & \textbf{Top-5}
& \textbf{Macro} & \textbf{Micro} & \textbf{Weighted} \\
\midrule
\multirow{4}{*}{\makecell{\textit{Individual-image}}}
& ResNet50
& \uline{65.5} & \uline{77.8} & \uline{81.9}
& \uline{0.33} & \uline{0.65} & \uline{0.62} \\
& DINOv3
& 63.6 & 76.4 & 81.0
& 0.31 & \uline{0.65} & 0.61 \\
& BioCLIPv2
& 52.8 & 66.1 & 70.8
& 0.21 & 0.51 & 0.47 \\
& \textbf{Pl@ntNet}
& \textbf{67.8} & \textbf{79.8} & \textbf{84.3}
& \textbf{0.37} & \textbf{0.69} & \textbf{0.66} \\
\midrule
\multirow{4}{*}{\makecell{\textit{Soft-voting}}}
& ResNet50
& 59.1 & 68.9 & 72.6
& 0.24 & 0.59 & 0.54 \\
& DINOv3
& \uline{72.3} & \uline{80.5} & \uline{82.4}
& \uline{0.39} & \uline{0.72} & \uline{0.68} \\
& BioCLIPv2
& 61.1 & 69.5 & 74.8
& 0.24 & 0.61 & 0.54 \\
& \textbf{Pl@ntNet}
& \textbf{74.3} & \textbf{81.6} & \textbf{83.7}
& \textbf{0.40} & \textbf{0.74} & \textbf{0.70} \\

\bottomrule
\end{tabular}
\end{table}

\begin{table}[t]
\centering
\caption{Performance on close-up images acquired on a single date.}
\label{tab:closeup}
\begin{tabular}{lcccccc}
\toprule
\textbf{Model} &
\multicolumn{3}{c}{\textbf{Accuracy (\%)}} &
\multicolumn{3}{c}{\textbf{F1 Score}} \\
\cmidrule(lr){2-4} \cmidrule(lr){5-7}
& \textbf{Top-1} & \textbf{Top-3} & \textbf{Top-5}
& \textbf{Macro} & \textbf{Micro} & \textbf{Weighted} \\
\midrule
ResNet50
& 40.8 & 55.1 & 63.9
& 0.14 & 0.42 & 0.36 \\
DINOv3
& \uline{76.8} & \textbf{86.9} & \textbf{89.1}
& \uline{0.39} & \uline{0.76} & \uline{0.71} \\
BioCLIPv2
& 59.5 & 66.3 & 67.3
& 0.22 & 0.59 & 0.54 \\
\textbf{Pl@ntNet}
& \textbf{77.9} & \uline{82.9} & \uline{83.9}
& \textbf{0.46} & \textbf{0.81} & \textbf{0.77} \\
\bottomrule
\end{tabular}
\end{table}

\begin{figure}[t]
\centering
\includegraphics[width=0.95\linewidth]{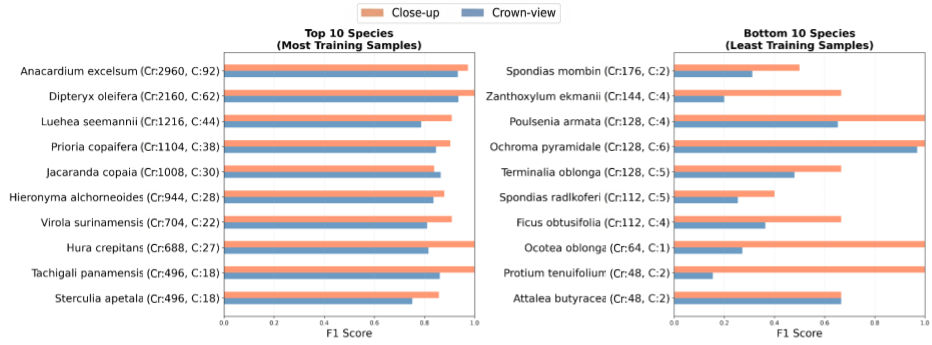}
\caption{F1 score comparison between crown-view (blue) and close-up (orange) 
models for top 10 (left) and bottom 10 (right) species by training sample 
size. Labels show training and test sample counts for crown-view (T) and close-up 
(C). Bottom 10 filtered for species with both non-zero F1 score with at least 1 training sample.}
\label{fig:scale_gap_results}
\end{figure}
We observe performance to be strongly dependent on the representation scale. For crown-view canopy imagery (Table~\ref{tab:topview}), soft voting consistently improves over per-image accuracy across all models. Among baselines, Pl@ntNet consistently performs best suggesting that representations learned from large-scale plant recognition data partially transfer to canopy imagery. In contrast, on close-up imagery (Table~\ref{tab:closeup}), both DINOv3 and Pl@ntNet show a large performance gain, with Pl@ntNet achieving the highest macro- and micro-F1 scores. 
As shown in Figure~\ref{fig:scale_gap_results}, for common species, both viewpoints achieve high and relatively stable performance, though close-up images consistently yield equal or higher F1 scores. For rare species, the gap becomes pronounced: crown-view performance degrades sharply with decreasing sample size, while close-up imagery maintains substantially higher F1 scores across most taxa, even in the extreme low-data regime (often fewer than 20 training instances). Notably, this advantage is achieved despite the close-up dataset containing far fewer total samples than the crown-view dataset.

\section{Discussion \& Future directions}
Our experiments reveal a persistent performance gap between canopy-level species classification from crown-view and close-up UAV imagery. While modern vision foundation models fine-tuned on segmented tree crown polygons achieve strong performance for common species, accuracy degrades substantially for rare taxa. This degradation is consistent across model families and is particularly pronounced in the long tail of the class distribution, where limited labeled data and subtle inter-species visual differences dominate. As future work, we will leverage unlabeled drone-acquired close-up imagery through a teacher–student representation transfer framework. A frozen Pl@ntNet model trained on close-up botanical images will serve as the teacher, while a Pl@ntNet-initialized student will be adapted to operate on crown-view canopy tiles. We will align embeddings via a cosine distillation loss to transfer species-relevant cues from close-up views to canopy-level representations. In conclusion, we systematically demonstrate a persistent representation gap between crown-view canopy imagery and drone based close-up photos for tropical tree species classification, with failures most pronounced under long-tailed, label-scarce regimes. Our results highlight the need for cross-scale representation alignment to transfer identifiable species cues into scalable canopy-level models.
\newpage
\bibliographystyle{iclr2026_conference}
\bibliography{references.bib}

\newpage
\appendix
\subsubsection*{Acknowledgments}
We would like to thank Alexis Joly, Jean-Christophe Lombardo, and Pierre Bonnet from the Pl@ntNet (\cite{lefort_cooperative_2026}) team for providing up-to-date weights for the pre-trained Pl@ntNet model and for their valuable insights. We are grateful to funding from the Canada CIFAR AI Chairs program, the Global Center on AI and Biodiversity Change (NSERC 585136), and the IVADO (R3AI, Postdoc Entrepreneur) program. This research was enabled in part by compute resources provided by Mila - Quebec AI Institute, including material support from NVIDIA Corporation.

\section{Appendix}

\subsection{Species Distribution Across Splits}
The following figures illustrate the species distribution across our training, validation, and test splits. Our dataset includes 84 species, with a significant portion of the distribution residing in the long tail; specifically, only 26 species have at least 20 labels.

\begin{figure}[htbp]
    \centering
    \includegraphics[width=\textwidth]{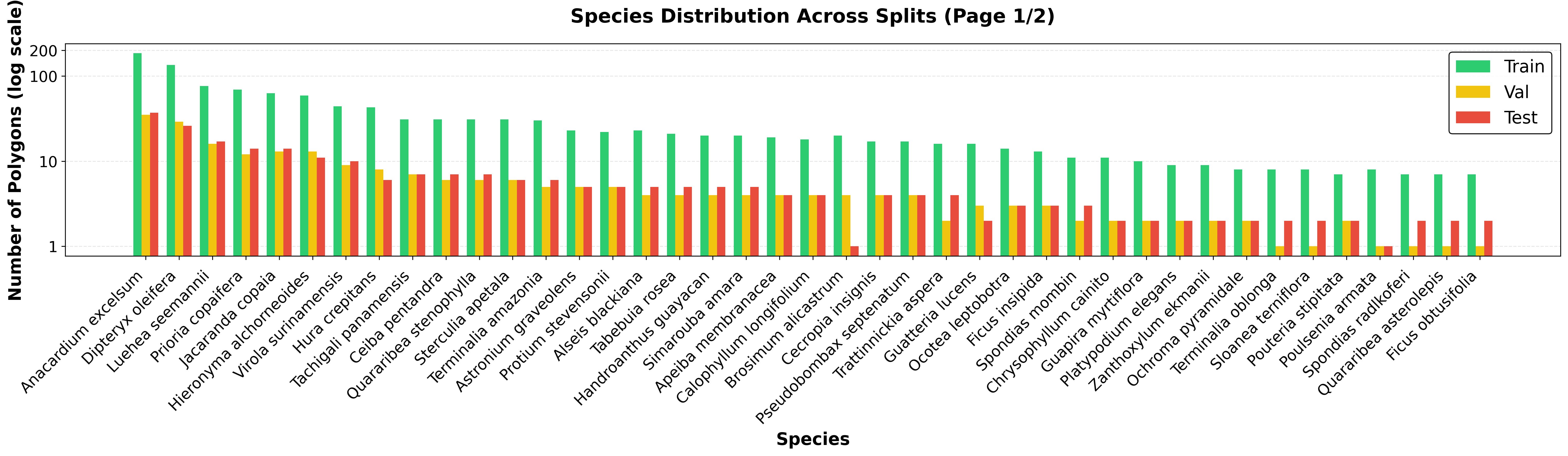}
    \caption{Species Distribution Across Splits (Page 1/2) showing the most common species in the dataset.}
    \label{fig:species_dist_1}
\end{figure}

\begin{figure}[htbp]
    \centering
    \includegraphics[width=\textwidth]{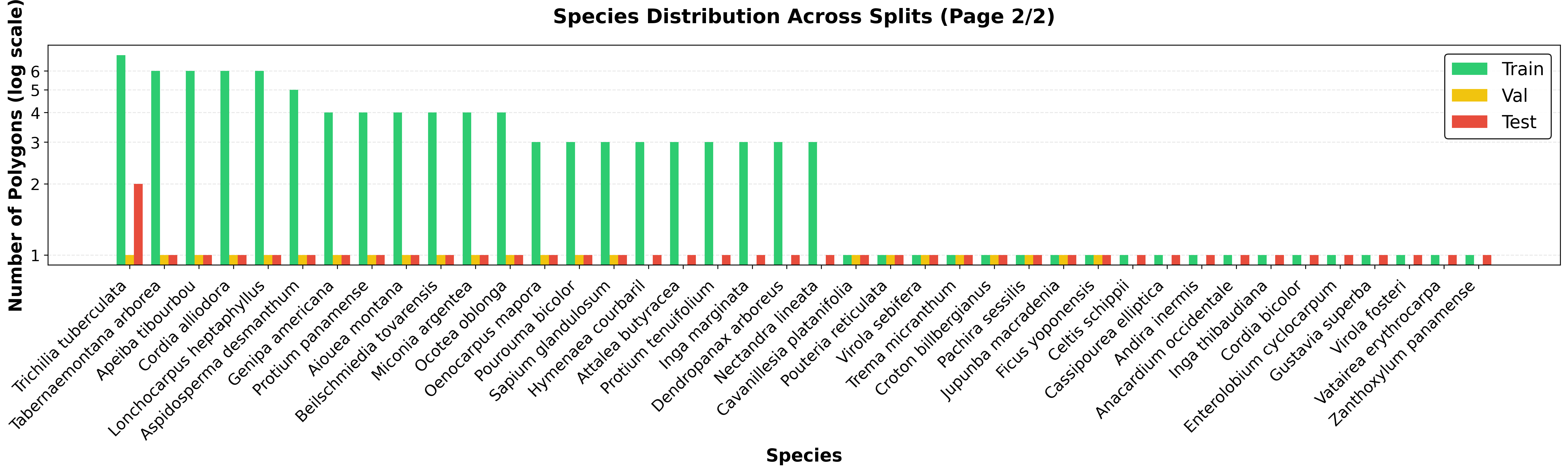}
    \caption{Species Distribution Across Splits (Page 2/2) highlighting the rare species in the long-tail distribution.}
    \label{fig:species_dist_2}
\end{figure}

\subsection{Temporal Variability}
\begin{figure}[htbp]
    \centering
    \begin{subfigure}[b]{0.48\textwidth}
        \centering
        \includegraphics[width=\textwidth]{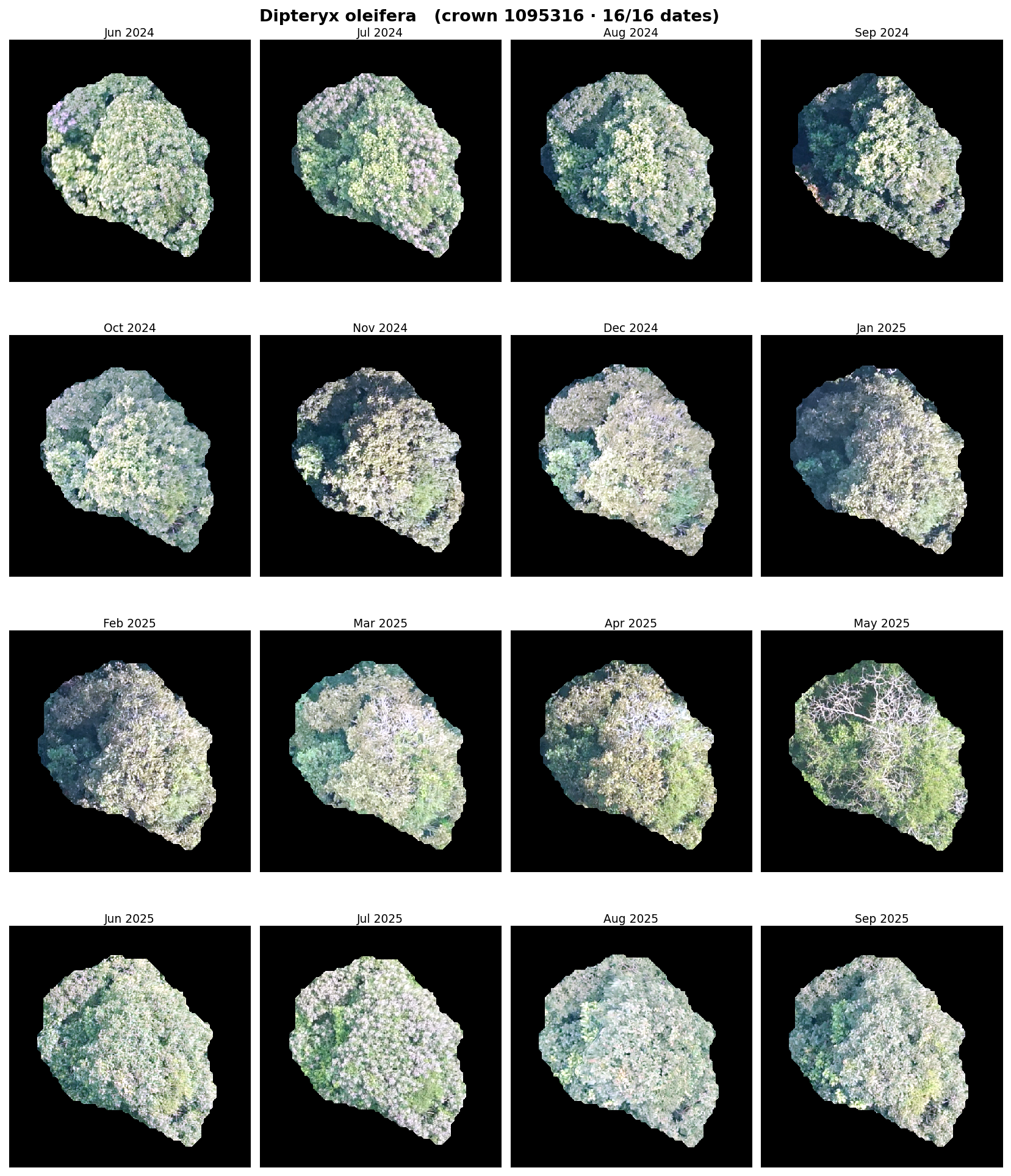}
        \caption{Commonly labeled species: Dipteryx Oleifera}
    \end{subfigure}
    \hfill
    \begin{subfigure}[b]{0.48\textwidth}
        \centering
        \includegraphics[width=\textwidth]{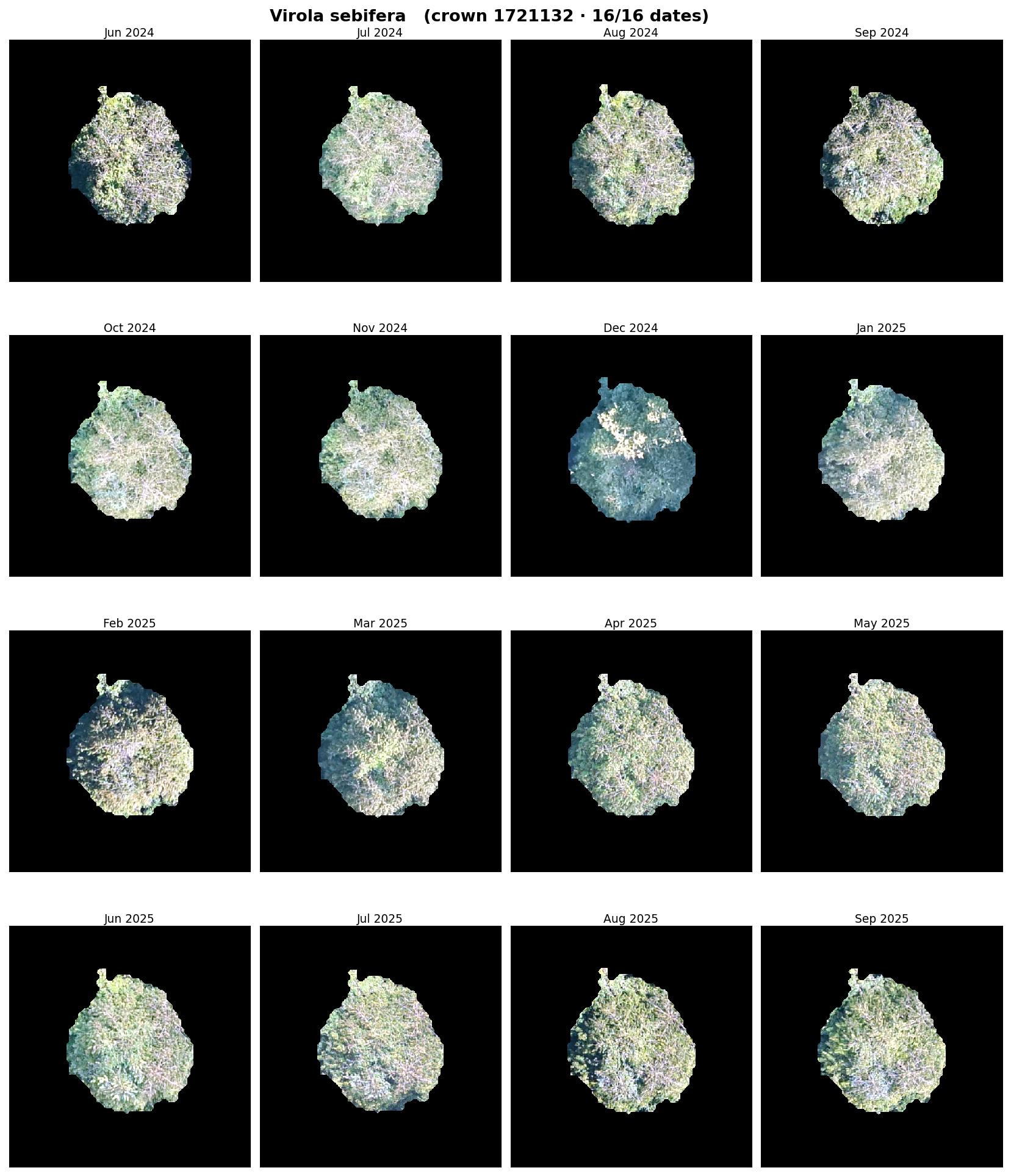}
        \caption{Rarely labeled species: Virola Sebifera}
    \end{subfigure}
    \caption{Observation of phenological variability over multiple months.}
    \label{fig:temporal_variability}
\end{figure}

The dataset time series captures phenological and illumination variability across up to 16 monthly snapshots. \ref{fig:temporal_variability} is a comparison of these temporal changes for a commonly labeled (Dipteryx Oleifera) and a rarely labeled species (Virola Sebifera).

\subsection{Model Specifications}
\ref{tab:model_specs} shows the model specifications. During training, we apply a standard set of geometric augmentations consistent across all models. Each image is first randomly cropped and resized to the target input resolution using RandomResizedCrop with a scale range of [0.7, 1.0] and bicubic interpolation, followed by a random rotation of up to ±30°. Random horizontal flipping (p=0.5) is also applied. No color or photometric augmentations are used. All images are normalized using ImageNet statistics (mean=[0.485, 0.456, 0.406], std=[0.229, 0.224, 0.225]).
\begin{table}[t]
\centering
\caption{%
  Model specifications and training hyperparameters for all fine-tuned backbones.
  All models use AdamW, mixed-precision training (fp16), batch size 32,
  and early stopping on validation loss (patience = 5, min-delta = 0.001)
  with a maximum of 100 epochs. Training time is total wall-clock time
  for 3-fold cross-validation on a single A100 GPU. Epochs are reported
  as mean\,$\pm$\,std across folds.
}
\label{tab:model_specs}
\setlength{\tabcolsep}{6pt}
\resizebox{\textwidth}{!}{%
\begin{tabular}{lcccc}
\toprule
 & \textbf{ResNet-50} & \textbf{DINOv3} & \textbf{BioCLIP-2} & \textbf{Pl@ntNet} \\
\midrule
\multicolumn{5}{l}{\textit{Architecture}} \\
\quad Backbone type        & CNN               & ViT-B/16          & ViT-B/16 (CLIP)    & ViT-B/14          \\
\quad Pretraining data     & ImageNet-1K       & LVD-1.68B         & Biological imagery & Citizen science plant imagery     \\
\quad Total parameters     & $\sim$25.6M       & $\sim$86M         & $\sim$149M         & $\sim$86M         \\
\quad Input resolution     & $224\times224$    & $512\times512$    & $224\times224$     & $518\times518$    \\
\quad Frozen components    & None              & None              & Text encoder       & None              \\
\midrule
\multicolumn{5}{l}{\textit{Hyperparameters}} \\
\quad Learning rate        & $1\times10^{-4}$  & $1\times10^{-4}$  & $5\times10^{-5}$   & $6\times10^{-6}$  \\
\quad Weight decay         & $1\times10^{-4}$               & $1\times10^{-4}$  & $0$                & $1\times10^{-4}$  \\
\quad Classifier dropout   & $0.0$             & $0.1$             & $0.0$              & $0.1$             \\
\midrule
\multicolumn{5}{l}{\textit{Training outcomes (3-fold cross-validation)}} \\
\quad Epochs trained       & $20 \pm 7$        & $16 \pm 4$        & $34 \pm 3$         & $18 \pm 4$        \\
\quad Total training time  & $\sim$49 min      & $\sim$2.2 h       & $\sim$9.6 h        & $\sim$8.1 h       \\
\bottomrule
\end{tabular}%
}
\end{table}

\subsection{Cross-Scale Visual Comparison}
\begin{figure}[htbp]
    \centering
    \includegraphics[width=\textwidth]{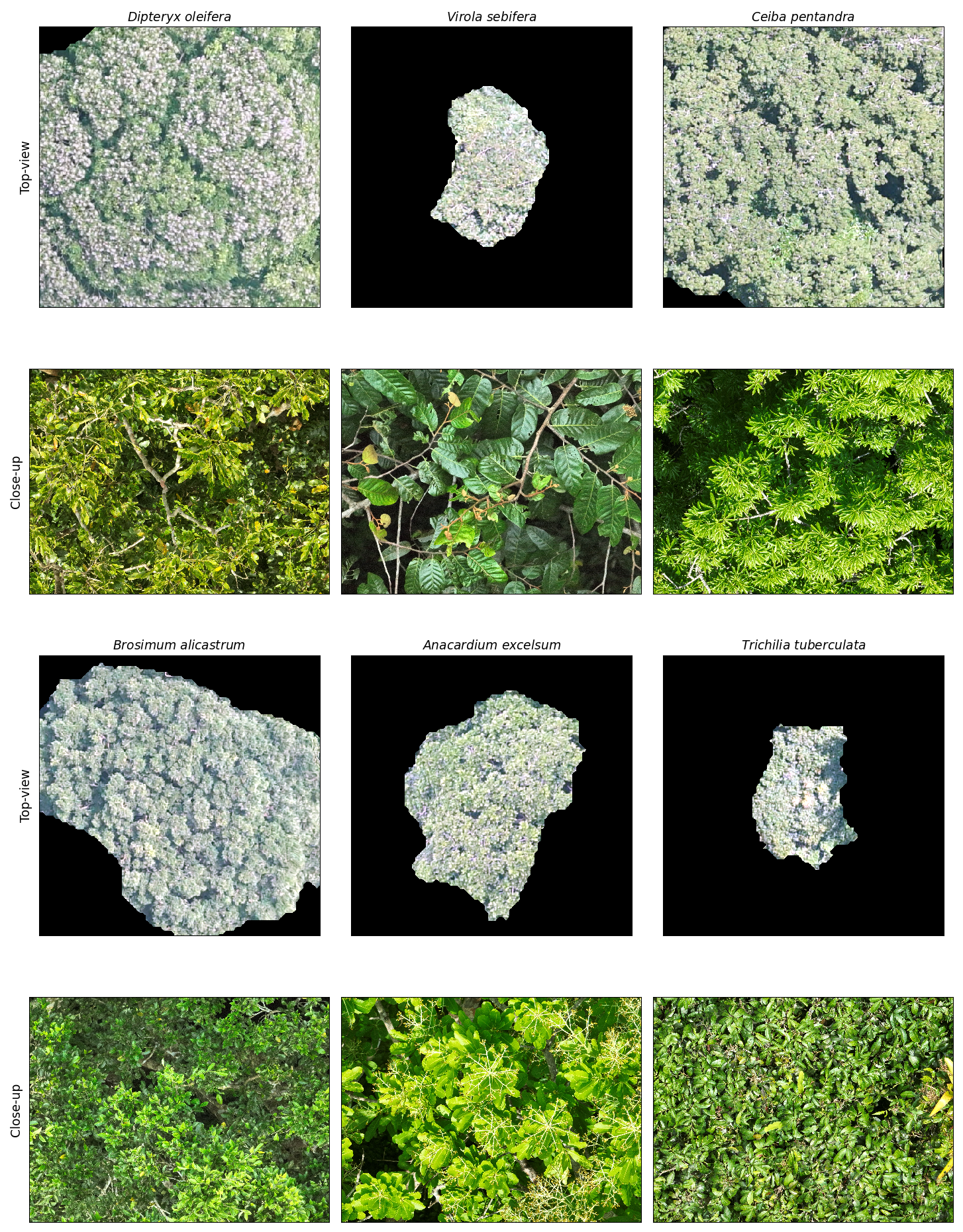}
    \hfill
    \caption{Paired crown-view and close-up drone imagery for six distinct species.}
    \label{fig:species_pairs}
\end{figure}

We provide visual examples of the two spatial scales used in our experiments: high-resolution close-up imagery (approx. 0.4 mm) and coarser-resolution top-view aerial imagery (4 cm GSD) \ref{fig:species_pairs}.

\section{Use of Large Language Models (LLMs)}
Large Language Models (LLMs) were utilized to assist with minor debugging of the LaTeX and Python code used in the experiments. The LLMs were not used for writing the manuscript, research ideation, data collection, or the generation of novel scientific conclusions. All conceptual, analytical, and experimental contributions remain the work of the authors.

\end{document}